\documentclass[letterpaper]{article}

\usepackage{aaai2027}

\usepackage[hyphens]{url}
\usepackage{graphicx}
\usepackage{natbib}
\usepackage{caption}
\usepackage{amsmath,amssymb,amsfonts}

\usepackage{booktabs}
\usepackage{multirow}
\usepackage{makecell}

\usepackage{xcolor}
\usepackage{colortbl}
\usepackage{pifont}

\title{CRISP: Constrained Refinement via Iterative Squeezing Process for Robust Medical Image Segmentation under Domain Shift}

\author{
Yizhou Fang\textsuperscript{\rm 1},
Pujin Cheng\textsuperscript{\rm 1,2},
Yixiang Liu\textsuperscript{\rm 1},
Xiaoying Tang\textsuperscript{\rm 1,3,*},
Longxi Zhou\textsuperscript{\rm 4,*}
}

\affiliations{
\textsuperscript{\rm 1}Department of Electronic and Electrical Engineering, Southern University of Science and Technology, Shenzhen, China\\
\textsuperscript{\rm 2}The University of Hong Kong, Hong Kong SAR, China\\
\textsuperscript{\rm 3}Jiaxing Research Institute, Southern University of Science and Technology, Jiaxing, China\\
\textsuperscript{\rm 4}Department of Biomedical Engineering, Southern University of Science and Technology, Shenzhen, China\\
\textsuperscript{*}Corresponding authors: tangxy@sustech.edu.cn, zhoulx@sustech.edu.cn
}

\copyrighttext{Preprint.}

\begin{document}

\maketitle

\begin{abstract}

Distribution shift in medical imaging remains a central bottleneck for the clinical translation of medical AI. Failure to address it can lead to severe performance degradation in unseen environments and exacerbate health inequities. Existing methods for domain adaptation are inherently limited by exhausting predefined possibilities through simulated shifts or pseudo-supervision. Such strategies struggle in the open-ended and unpredictable real world, where distribution shifts are effectively infinite. To address this challenge, we adopt the “Rank Stability of Positive Regions” as a working assumption under distribution shift, and use it to derive robust spatial hints for source-only segmentation. Guided by this assumption, we propose CRISP, a model-agnostic framework that, unlike deployment-time adaptation, requires no test-time parameter updates and no target-domain data---a target-free, plug-in refinement framework that segments with frozen weights. Rather than using ranking to directly output masks, CRISP exploits the stability of probability rankings under distribution shift to derive robust spatial priors. Via latent feature perturbation, perturbation-invariant high-grade regions define a high-precision (HP) core, while voxels that remain potentially foreground under at least one perturbation define a high-recall (HR) support; these dual priors are then recursively refined under perturbation. We then design an iterative training framework that progressively squeezes HP and HR toward the final segmentation. Extensive evaluations on multi-center cardiac MRI and CT-based lung vessel segmentation demonstrate CRISP's superior robustness, significantly outperforming state-of-the-art methods with striking HD95 reductions of up to 0.14 (7.0\% improvement), 1.90 (13.1\% improvement), and 8.39 (38.9\% improvement) pixels across multi-center, demographic, and modality shifts, respectively.
\end{abstract}


\section{Introduction}

\begin{figure*}[!t]
\centering
\includegraphics[width=1.0\textwidth]{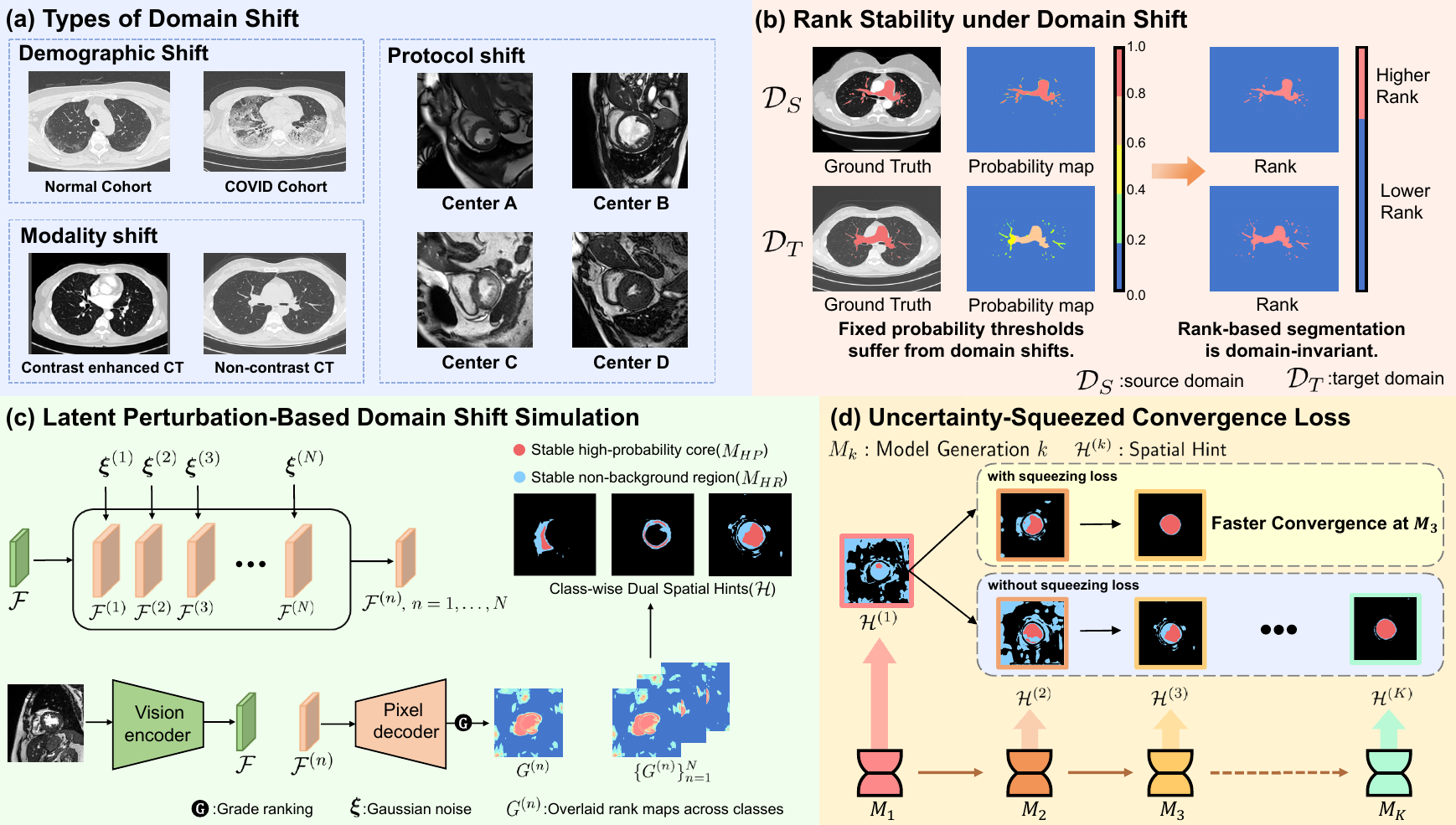}
\caption{Overview of CRISP, a rank-based segmentation framework rather than probability-driven prediction: predict $\rightarrow$ estimate rank-stability $\rightarrow$ get HP and HR $\rightarrow$ predict again with HP and HR $\rightarrow$ estimate rank-stability again $\rightarrow$ better HP and HR $\rightarrow \dots \rightarrow$ HP and HR converge to final segmentation. (a) Three representative clinical shifts: demographic, modality, and protocol variations. (b) Empirical Rank Stability: the relative rank of foreground regions remains stable despite probability drifts across domains. (c) Latent Feature Perturbation: injecting Gaussian noise into bottleneck features to simulate domain shift, yielding stochastic rank maps. Across perturbations, consistently top-graded voxels form stable high-probability ($M_{HP}$) core, while persistently non-background voxels form the stable non-background region ($M_{HR}$). (d) Recursive Self-Evolution: guided by these hints, successive model generations progressively squeeze the ambiguity region between $M_{HP}$ and $M_{HR}$, progressively bringing the two boundaries closer until convergence forms the final segmentation.}
\label{fig:1}
\end{figure*}

Deep learning has demonstrated state-of-the-art performance in medical image segmentation, crucial for computer-aided diagnosis~\cite{chu2026horuseye,zhou2020rapid,litjens2017survey,ronneberger2015u,jiangtao2025comprehensive}. However, its widespread clinical utility is constrained by distribution shift: models trained on source databases often suffer from significant performance degradation when deployed to unseen clinical environments, undermining confidence in clinical utility~\cite{lawrence2025artificial,gu2021domain}. This problem further widens the social gap, as models optimized for developed nations fail to accommodate the racial and geographic diversity of developing regions. Practically, as shown in Fig.~\ref{fig:1}(a), distribution shift can be classified as~\cite{yoon2024domain}: 1) Demographic Shift, arising from patient heterogeneity in age, gender, and disease distribution; and 2) Protocol and Modality Shift, driven by acquisition diversity such as variable imaging protocols, noise, and resolution, which collectively hinder cross-site transferability.

To address distribution shift, existing research can be broadly categorized into two paradigms: Training-time Generalization, which focuses on learning invariant representations~\cite{cheng2025dynamic,liu2021feddg,liu2025spectrum,carlucci2019domain}, and Deployment-time Adaptation, which dynamically updates model parameters during inference~\cite{zhang2024iplc,zhou2025tegda,chen2025dual,nado2020evaluating}. However, we argue that these approaches are fundamentally limited by a methodology of exhaustive adaptation. They attempt to either simulate an infinite space of shifts or reactively tune parameters, but both strategies struggle in the unpredictable real world. In contrast, our framework builds on a simple assumption---the ``Rank Stability of Positive Regions''---that the relative rank of positive predictions remains largely stable under distribution shifts. As shown in Fig.~\ref{fig:1}(b), although the predicted probability of a lesion may change substantially (for example, from 0.9 to 0.4) under distribution shift, the relative rank ordering of the true positive region remains largely preserved, enabling rank-based segmentation to correctly retain the lesion even under severe distribution shift.

This ``Rank Stability'' assumption is grounded in both empirical evidence and a decision-boundary interpretation. Empirically, DLPE~\cite{zhou2022interpretable} assumes that when segmenting blood vessels and airways, positive voxels always rank higher than negative voxels under domain shifts. Based on this assumption, DLPE robustly extracts blood vessels and airways under strong clinical shifts by ranking response maps and retaining the top-rated voxels according to a predefined anatomical volume ratio (e.g., pulmonary vessels are predefined to occupy 7.5\% of the lung volume). This suggests that relative ranking between positives and negatives largely remains stable even under strong domain shifts. However, DLPE handles only a predefined anatomical volume ratio, whereas our proposed CRISP handles general segmentation tasks.

A decision-boundary perspective offers further intuition for the ``Rank Stability'' assumption. Intuitively, voxels far from the decision boundary retain a stable rank across perturbations, whereas boundary voxels are rank-unstable; this motivates deriving reliable spatial guidance from the rank-consistent regions while leaving the ambiguous boundary to be refined. This perspective has also inspired rudimentary applications in classification tasks. Notably, it has been demonstrated that preserving the relative order of predictions, rather than enforcing absolute consistency, significantly enhances robustness against domain shifts~\cite{jing2023order}; furthermore, even when deep models exhibit miscalibration in complex environments, the ranking of class labels has proven to be the most reliable discriminative information~\cite{luo2024trustworthy}. However, these works are limited to sample-level class rankings and lack the systematic formulation required for pixel-level dense prediction. Even RankSEG~\cite{dai2023rankseg}, the existing ranking-based segmentation rule, employs probability ranking and volume estimation in a fully supervised setting and does not target distribution shift; the potential of rank stability under domain shift in segmentation thus remains largely unexplored. We therefore adopt Rank Stability as a working assumption for framework design: CRISP does not use ranking to output masks, but instead exploits perturbation-induced rank stability to derive high-precision (HP) and high-recall (HR) hints for source-only domain generalization.

In summary, this paper makes the following key contributions to the field of medical image segmentation:
\begin{itemize}
\setlength{\itemsep}{2pt}
\item We adopt the ``Rank Stability of Positive Regions'' as a critical assumption under distribution shift, serving as a reliable anchor against open-world distribution shift.
\item We propose CRISP, a model-agnostic framework that requires no parameter updates at inference (operating with frozen weights) and no target-domain data, transforming this assumption into an actionable strategy driven by rank stability rather than absolute probabilities; it iteratively squeezes the uncertainty gap between High-Precision and High-Recall to form the final segmentation.
\item Extensive evaluations on multi-center cardiac MRI and CT-based lung vessel segmentation show CRISP's superior robustness, significantly outperforming state-of-the-art methods with HD95 reductions of up to 0.14~(7.0\%), 1.90~(13.1\%), and 8.39~(38.9\%) pixels across multi-center, demographic, and modality shifts, respectively.
\end{itemize}

\section{Method}
\noindent\textbf{Overview.} Our method updates the high-precision (HP) and high-recall (HR) hints step by step, and the two progressively converge to the final segmentation. As illustrated in Fig.~\ref{fig:1}, the process is a simple loop: predict $\rightarrow$ estimate rank-stability $\rightarrow$ get HP and HR $\rightarrow$ predict again with HP and HR $\rightarrow$ estimate rank-stability again $\rightarrow$ better HP and HR $\rightarrow \cdots \rightarrow$ HP and HR converge to the final segmentation.

\noindent\textbf{Problem Definition.}
We study source-only domain generalization for volumetric medical image segmentation. Given a labeled source domain $\mathcal{D}_S=\{(X_S,Y_S)\}$ with input volumes $X\in\mathbb{R}^{H\times W\times D}$ and voxel-wise labels $Y\in\{0,1,\dots,C\}^{H\times W\times D}$ over $C$ foreground classes, our goal is to generalize to unseen domains experiencing distribution shifts. To this end, we propose the CRISP framework (Fig.~\ref{fig:1}). Grounded in the Rank Stability assumption, CRISP simulates domain shift via Latent Feature Perturbation to extract domain-invariant spatial hints: a stable high-probability core ($M_{HP}$) and a stable non-background region ($M_{HR}$). These dual hints drive a Recursive Self-Evolution process, progressively squeezing the uncertainty gap for precise, target-free boundary refinement.

Although the task and all evaluation metrics are volumetric, CRISP is implemented slice-by-slice: each 2D axial slice $X_{sl}$ is processed independently and the per-slice predictions are stacked to reconstruct the 3D volume. This design enables fair comparison with MedSAM-based~\cite{ma2024segment} baselines such as IPLC~\cite{zhang2024iplc}, which are inherently restricted to 2D slices. Throughout, $X$ denotes a volume and $X_{sl}$ one of its slices.

\subsection{Recursive Self-Evolution Training Framework}
\label{sec:2.1}
\noindent\textbf{Hint-conditioned input.} CRISP conditions a standard segmentation network on spatial hints by channel concatenation. For a task with $C$ foreground classes, the input stacks the image slice with per-class hint maps: one image channel, $C$ high-precision channels $M_{HP}$, and $C$ high-recall channels $M_{HR}$ (e.g., $1{+}3{+}3{=}7$ channels for the LV/MYO/RV cardiac task). The hint pair at generation $k$ is denoted $\mathcal{H}^{(k)}=\{M_{HP}^{(k)},M_{HR}^{(k)}\}$. The only architectural change is widening the first convolution to accept these $2C$ extra channels; no additional trainable modules are introduced.

\noindent\textbf{Generational training and offline hint generation.} The first model $f_{\theta_1}$ is trained from trivial hints---an empty core $M_{HP}^{(1)}=\mathbf{0}$ and a full-image support $M_{HR}^{(1)}=\mathbf{1}$---so that it learns to segment without spatial guidance. Once $f_{\theta_{k-1}}$ has converged, we freeze its best weights and generate tighter hints offline: starting from the trivial hints, the perturbation-and-ranking procedure (Sec.~\ref{sec:2.2}) is applied $d$ times in sequence to produce a depth-$d$ hint pair, where a larger $d$ yields a tighter bracket around the target. Each subsequent $f_{\theta_k}$ ($k\geq2$) is then fine-tuned from $f_{\theta_{k-1}}$ on the expanded dataset described next.

\noindent\textbf{Balanced dataset expansion.} A key design choice is that generation $k$ is trained not on a single hint depth but on a balanced mixture of all depths produced so far. Concretely, $D_1$ contains only depth-$0$ (trivial) hints; $D_2$ mixes $50\%$ depth-$0$ and $50\%$ depth-$1$ hints (from $f_{\theta_1}$); $D_3$ mixes $33\%$ each of depth-$0$, depth-$1$, and depth-$2$ hints (from $f_{\theta_2}$); and in general $D_k$ draws depths $0,1,\dots,k-1$ in equal proportion. Exposing one network to the full spectrum of hint tightness---from coarse to refined---lets a single model iteratively refine its own hints regardless of their current quality, which is precisely what makes recurrent single-model inference possible.

\noindent\textbf{Convergence.} By construction, the rank-stable core lies inside the foreground, which lies inside the rank-stable support, $M_{HP}^{(k)}\subseteq\hat{Y}\subseteq M_{HR}^{(k)}$, so the uncertainty band $\Omega_u^{(k)}=M_{HR}^{(k)}\setminus M_{HP}^{(k)}$ brackets the true boundary. Each generation shrinks this band, driving both bounds toward the target, i.e., $\lim_{k \to \infty} M_{HP}^{(k)} = \lim_{k \to \infty} M_{HR}^{(k)} = \hat{Y}$.

\noindent\textbf{Single-model recurrent inference.} At test time we deploy only the final model $f_{\theta_K}$ with frozen weights. Starting from trivial hints $(\mathbf{0},\mathbf{1})$, we apply the perturbation-and-ranking procedure to progressively tighten the HP/HR pair, and after these refinements $f_{\theta_K}$ produces the final prediction from the converged hints (the perturbation-based hint refinement and the final clean prediction are separate passes). Thus $f_{\theta_1}$ predicts directly from trivial hints, whereas the deployed $f_{\theta_K}$ refines its hints before the final read-out; the entire recurrence reuses one set of weights and requires neither test-time parameter updates nor target-domain data.

\subsection{Latent Feature Perturbation for Constraint Generation}
\label{sec:2.2}
We employ Latent Feature Perturbation to simulate domain shifts under the Rank Stability assumption, assuming that genuine structures largely preserve their relative probability ordering across shifts. We emphasize that Gaussian perturbation is not meant to reproduce every real-world shift; it probes rank stability by displacing voxels in proportion to their proximity to the decision boundary, so that rank-stable voxels can be separated from rank-unstable boundary voxels. Alternative perturbations (image-space, Fourier, SNR, or other feature-space noise) could serve the same purpose, and latent Gaussian noise is adopted as a simple, model-agnostic instantiation. That this perturbation-probed stability transfers to genuine shifts is borne out by CRISP's consistent gains across all three real-world shifts---multi-center, modality, and demographic.

Given an input slice $X_{sl}$ and spatial hints $\mathcal{H}^{(k)}$ (concatenated along the channel dimension), the encoder $f_{\text{enc}}$ extracts latent features
\begin{equation}
\mathcal{F} = f_{\text{enc}}(X_{sl}, \mathcal{H}^{(k)}),
\end{equation}
which form the deep bottleneck, where domain-specific appearance (acquisition ``style'') is most compactly encoded; perturbing here therefore emulates a change of domain far more efficiently than perturbing raw pixels. To simulate such shifts, we draw $N$ independent Gaussian noise samples $\xi^{(n)} \sim \mathcal{N}(0, \sigma^2 I)$ and inject each into the bottleneck:
\begin{equation}
\mathcal{F}^{(n)} = \mathcal{F} + \xi^{(n)}, \quad n=1,\dots,N.
\end{equation}
Here, $\sigma$ controls the severity of the simulated shift and the $N$ draws together emulate an ensemble of neighbouring domains. Decoding each perturbed feature, $f_{\text{dec}}$ produces a slightly different pixel-wise probability map
\begin{equation}
P^{(n)} = \mathrm{Softmax}(f_{\text{dec}}(\mathcal{F}^{(n)})),
\end{equation}
where $P^{(n)}_{c}(x,y)$ denotes the class-$c$ probability at pixel $(x,y)$; comparing these $N$ maps reveals which voxels keep their relative ordering and which do not.

Probabilities saturate near $0$ and $1$, where large evidence differences are compressed into negligible numerical gaps, making rank comparisons unreliable. To map the predictions into a space that instead reflects the accumulated amount of evidence linearly and without bounds, we transform the probabilities to log-odds
\begin{equation}
S_c^{(n)} = \ln \frac{P_c^{(n)}}{1-P_c^{(n)}}.
\end{equation}
Because the absolute log-odds scale varies from slice to slice (and from domain to domain), we normalize using instance-level extremes so that grades are comparable across images:
\begin{equation}
S_{\min}^{(n)} = \min_{c,x,y} S_c^{(n)}(x,y),
\end{equation}
\begin{equation}
S_{\max}^{(n)} = \max_{c,x,y} S_c^{(n)}(x,y).
\end{equation}
The grade map is defined as
\begin{equation}
G_c^{(n)} =
\left\lfloor
\frac{S_c^{(n)} - S_{\min}^{(n)}}{S_{\max}^{(n)} - S_{\min}^{(n)}} \times L
\right\rfloor_{\text{clamped}},
\end{equation}
with $G_c^{(n)}(x,y) \in \{0,\dots,L-1\}$, where $\lfloor\cdot\rfloor_{\text{clamped}}$ clips values into $\{0,\dots,L-1\}$. Higher grades indicate stronger foreground confidence. Quantizing the continuous log-odds into $L$ coarse grades is deliberate: it discards small numerical jitter and judges stability at the level of rank bands rather than exact probabilities, so a voxel only changes grade when its evidence genuinely shifts.

From the $N$ stochastic grade maps, we derive the class-wise masks $M_{HP,c}^{(k)}$ and $M_{HR,c}^{(k)}$ for each class $c$:
\begin{equation}
M_{HP,c}^{(k)}
=
\bigcap_{n=1}^{N}
\{G_c^{(n)} = L-1\},
\end{equation}
\begin{equation}
M_{HR,c}^{(k)}
=
\bigcup_{n=1}^{N}
\{G_c^{(n)} > 0\}.
\end{equation}
The two masks use complementary set operations. $M_{HP,c}^{(k)}$ is an intersection: a voxel is kept only if it attains the top grade under every one of the $N$ perturbations, so it survives all simulated shifts---a rank-stable, high-precision core that is almost surely foreground. $M_{HR,c}^{(k)}$ is a union: a voxel is kept if it is non-background under any perturbation, yielding an inclusive, high-recall support that almost surely contains the entire foreground. The uncertainty region is their difference,
\begin{equation}
\Omega_{u,c}^{(k)} = M_{HR,c}^{(k)} \setminus M_{HP,c}^{(k)},
\end{equation}
i.e., exactly the rank-unstable voxels whose grade flips across perturbations---the fragile boundary that the next generation must resolve.

\begin{table*}[!t]
\centering
\caption{Quantitative comparison on M\&Ms dataset. Dice score (\%) $\uparrow$ and HD95 (pixel) $\downarrow$. \textbf{Bold} = best among all domain-generalization methods; \colorbox{red!25}{red} = cells where our method ranks first among source-only (\textcolor{green}{$^{\star}$}) methods, the strictly like-for-like comparison. \textcolor{green}{$^{\star}$} = source-only methods that cannot access the target domain. \textcolor{red}{$^{\#}$} = target-informed methods that require target-domain data to update parameters (need retraining). \textcolor{red}{\textbf{\textcolor{red}{$^{\#}$} and \textcolor{green}{$^{\star}$} are not a fair comparison, as \textcolor{red}{$^{\#}$} requires more information (information from the target domain) than \textcolor{green}{$^{\star}$}.}} \textbf{A+T Ref.} = joint supervised training on labeled source $A$ and one labeled target domain---a strong reference, not an oracle ceiling.}
\label{tab:results}
\renewcommand{\arraystretch}{1.3}
\resizebox{\textwidth}{!}{
\begin{tabular}{l|l|ccc|ccc|ccc}
\hline
\multirow{2}{*}{Metrics} & \multirow{2}{*}{Method} & \multicolumn{3}{c|}{Target domain B} & \multicolumn{3}{c|}{Target domain C} & \multicolumn{3}{c}{Target domain D} \\ \cline{3-11}
 &  & LV & MYO & RV & LV & MYO & RV & LV & MYO & RV \\ \hline
\multirow{7}{*}{\begin{tabular}[c]{@{}l@{}}Dice $\uparrow$\\ (\%)\end{tabular}}
 & Baseline & 85.22$\pm$13.27 & 75.91$\pm$11.22 & 78.25$\pm$18.15 & 83.32$\pm$11.86 & 74.54$\pm$9.15 & 78.30$\pm$14.84 & 88.36$\pm$8.22 & 75.76$\pm$5.44 & 83.57$\pm$10.21 \\
 & A+T Ref. &90.92$\pm$7.09 & 85.10$\pm$3.73 & 86.47$\pm$9.39 & 88.60$\pm$8.58 & 81.54$\pm$7.12 & 85.78$\pm$7.12 & 92.75$\pm$4.79 & 81.67$\pm$3.30 & 87.56$\pm$17.83 \\ \cline{2-11}
 & \textcolor{green}{FedDG$^{\star}$} & 86.97$\pm$12.18 & 79.25$\pm$5.88 & 85.05$\pm$9.72 & 87.17$\pm$7.99 & 76.96$\pm$7.65 & 83.65$\pm$10.20 & 89.91$\pm$5.86 & 76.83$\pm$4.26 & 86.89$\pm$6.71 \\
 & \textcolor{green}{DDG-Med$^{\star}$} & 89.01$\pm$9.78 & 80.15$\pm$5.08 & 84.81$\pm$9.51 & 88.67$\pm$6.70 & 78.62$\pm$6.39 & 83.33$\pm$8.55 & 90.75$\pm$3.27 & 78.40$\pm$4.94 & 85.34$\pm$7.11 \\
 & \textcolor{red}{IPLC$^{\#}$} & 88.50$\pm$6.23 & \textbf{83.27$\pm$3.70} & 83.40$\pm$9.01 & 88.62$\pm$9.32 & 78.82$\pm$7.40 & 84.08$\pm$14.39 & 89.91$\pm$5.63 & 79.24$\pm$3.64 & 86.28$\pm$6.99 \\
 & \textcolor{red}{TEGDA$^{\#}$} & 88.98$\pm$9.28 & 83.21$\pm$3.43 & 85.52$\pm$7.60 & 85.57$\pm$9.64 & 78.01$\pm$7.24 & 82.90$\pm$10.71 & 89.58$\pm$7.10 & 79.58$\pm$3.68 & \textbf{87.46$\pm$4.92} \\
 & \textcolor{green}{Ours$^{\star}$} & \cellcolor{red!25}\textbf{89.57$\pm$8.55} & \cellcolor{red!25}82.74$\pm$4.19 & \cellcolor{red!25}\textbf{86.49$\pm$8.53} & \cellcolor{red!25}\textbf{88.92$\pm$6.32} & \cellcolor{red!25}\textbf{79.82$\pm$8.41} & \cellcolor{red!25}\textbf{84.46$\pm$10.03} & \cellcolor{red!25}\textbf{92.28$\pm$3.04} & \cellcolor{red!25}\textbf{80.81$\pm$4.51} & 85.26$\pm$7.61 \\ \hline
\multirow{7}{*}{\begin{tabular}[c]{@{}l@{}}HD95 $\downarrow$\\ (pixel)\end{tabular}}
 & Baseline & 2.64$\pm$2.68 & 2.15$\pm$1.22 & 3.80$\pm$5.88 & 3.57$\pm$4.57 & 2.66$\pm$1.87 & 3.94$\pm$5.90 & 2.25$\pm$2.75 & 2.05$\pm$0.81 & 2.55$\pm$2.87 \\
 & A+T Ref. &1.72$\pm$1.47 & 1.55$\pm$0.62 & 2.08$\pm$1.64 & 2.84$\pm$4.04 & 2.39$\pm$2.37 & 2.42$\pm$1.54 & 2.92$\pm$4.77 & 1.45$\pm$0.48 & 1.48$\pm$1.28 \\ \cline{2-11}
 & \textcolor{green}{FedDG$^{\star}$} & 2.47$\pm$2.03 & 2.17$\pm$1.05 & 3.65$\pm$5.53 & 2.96$\pm$3.05 & 2.32$\pm$0.95 & 3.11$\pm$2.45 & 2.53$\pm$3.44 & 2.12$\pm$0.83 & 2.98$\pm$5.47 \\
 & \textcolor{green}{DDG-Med$^{\star}$} & 2.43$\pm$3.38 & 2.23$\pm$1.16 & 3.03$\pm$2.83 & 2.29$\pm$1.62 & \textbf{2.04$\pm$0.80} & 3.05$\pm$2.46 & \textbf{1.78$\pm$0.59} & 2.30$\pm$0.92 & 3.65$\pm$3.54 \\
 & \textcolor{red}{IPLC$^{\#}$} & \textbf{1.74$\pm$1.25} & 1.86$\pm$1.17 & 2.05$\pm$1.46 & 2.97$\pm$3.40 & 2.17$\pm$1.39 & 2.92$\pm$3.15 & 1.79$\pm$2.01 & 1.80$\pm$0.52 & 2.04$\pm$1.82 \\
 & \textcolor{red}{TEGDA$^{\#}$} & 1.88$\pm$1.99 & \textbf{1.53$\pm$0.56} & 1.86$\pm$1.17 & 2.65$\pm$2.43 & 2.49$\pm$1.74 & 2.40$\pm$2.28 & 2.10$\pm$3.20 & 1.66$\pm$0.45 & 1.44$\pm$0.52 \\
 & \textcolor{green}{Ours$^{\star}$} & \cellcolor{red!25}1.80$\pm$1.30 & \cellcolor{red!25}1.57$\pm$0.64 & \cellcolor{red!25}\textbf{1.57$\pm$1.21} & \cellcolor{red!25}\textbf{2.06$\pm$1.63} & 2.17$\pm$1.60 & \cellcolor{red!25}\textbf{2.20$\pm$3.05} & 2.43$\pm$4.57 & \cellcolor{red!25}\textbf{1.60$\pm$0.53} & \cellcolor{red!25}\textbf{1.38$\pm$0.86} \\ \hline
\end{tabular}
}
\end{table*}

\subsection{Accelerated Convergence: Uncertainty Squeezing Loss}
\label{sec:2.3}
To mitigate boundary ambiguity between $M_{HR,c}^{(k)}$ and $M_{HP,c}^{(k)}$ and accelerate refinement, we introduce an Uncertainty Squeezing Loss ($\mathcal{L}_{\text{squeeze}}$) to enforce predictive consistency within the global uncertainty region $\Omega_u^{(k)} = \bigcup_{c=1}^{C} \Omega_{u,c}^{(k)}$:
\begin{equation}
\mathcal{L}_{\text{squeeze}} = \frac{1}{|\Omega_u^{(k)}| + \epsilon} \sum_{n=1}^{N} \left\| \left( P_k - P_k^{(n)} \right) \odot \Omega_u^{(k)} \right\|^2,
\end{equation}
where $P_k$ and $P_k^{(n)}$ denote the predictions at generation $k$ and under the $n$-th perturbation, respectively, and $\odot$ is element-wise multiplication.

Both the restriction to $\Omega_u^{(k)}$ and the normalization by its area are deliberate. By construction, the high-precision core $M_{HP}^{(k)}$ and the background complement $1-M_{HR}^{(k)}$ are rank-stable---their predictions barely vary across perturbations and are already (near-)certain---so the only genuinely uncertain, perturbation-sensitive region, and the only region the model is left free to explore, is the band $\Omega_u^{(k)}$ itself. Confining the penalty to $\Omega_u^{(k)}$ therefore measures instability only where it matters, while dividing by its area $|\Omega_u^{(k)}|$ keeps the loss on a comparable scale even as the band shrinks across generations.

This also makes $\mathcal{L}_{\text{squeeze}}$ behave as a regularizer whose effective strength adapts to the current band width: as the gap narrows, enforcing perturbation-invariance over an ever-thinner region pushes the network toward increasingly decisive predictions that commit each ambiguous voxel rather than hedging. Such behavior is precisely what convergence requires, because near convergence $M_{HP}^{(k)}$ and $M_{HR}^{(k)}$ are close and the residual positive evidence inside $\Omega_u^{(k)}$ becomes extremely weak---so weak that a small perturbation can scramble its rank, exactly where the rank-stability signal is most fragile. By enforcing perturbation-invariance in this thin band, $\mathcal{L}_{\text{squeeze}}$ teaches the model to respond stably and decisively when HP and HR are close, allowing the two boundaries to keep squeezing toward $\hat{Y}$ instead of stalling.

The overall objective combines standard Dice loss with the proposed squeezing loss:
\begin{equation}
\mathcal{L}_{\text{total}} = \mathcal{L}_{\text{dice}} + \alpha_d\, \mathcal{L}_{\text{squeeze}},
\end{equation}
where $\mathcal{L}_{\text{dice}}$ is the Dice loss between the prediction $P_k$ and the target label, and $\alpha_d$ is a depth-dependent weight applied per training sample:
\begin{equation}
\alpha_d =
\begin{cases}
0, & d=0,\\
\alpha, & d>0,
\end{cases}
\end{equation}
with $\alpha>0$ a fixed hyperparameter. The squeezing loss is thus disabled for depth-$0$ samples---whose trivial hints make $\Omega_u$ the whole image with no useful structure---and active only once meaningful HP/HR hints exist ($d>0$). This is consistent across generations and also applies to the first model $f_{\theta_1}$, which sees only depth-$0$ hints and is therefore trained with Dice loss alone.

\begin{table*}[!t]
\centering
\caption{Quantitative evaluation across different clinical scenarios. \textbf{(S)}: Source domain, \textbf{(T)}: Target domain. Dice score (\%) $\uparrow$ and HD95 (pixel) $\downarrow$. \textbf{Bold} = best result. \textcolor{green}{$^{\star}$} = Methods that cannot access the target domain. \textcolor{red}{$^{\#}$} = Methods that need the target domain to update parameters (need more information, need retraining). \textcolor{red}{\textbf{Remarkably, our source-only (\textcolor{green}{$^{\star}$}) method even outperforms the deployment-time adaptation methods (\textcolor{red}{$^{\#}$}), despite using no target-domain data.}} Source-domain (S) columns are reported to verify source-performance preservation; the target-adaptive methods ($\#$) adapt only on the target split, so their source scores coincide with the source model.}
\label{tab:combined_scenarios}
\renewcommand{\arraystretch}{1}
\resizebox{1.0\textwidth}{!}{
\begin{tabular}{l|l|cc|cc}
\hline
\multirow{2}{*}{Metrics} & \multirow{2}{*}{Method} & \multicolumn{2}{c|}{Scenario 1 (Demographic)} & \multicolumn{2}{c}{Scenario 2 (Modality)} \\ \cline{3-6}
 &  & Normal Cohort (S) & COVID Cohort (T) & Contrast-enhanced (S) & Non-contrast (T) \\ \hline
\multirow{4}{*}{\begin{tabular}[c]{@{}l@{}}Dice $\uparrow$\\ (\%)\end{tabular}}
 & Baseline & 81.14$\pm$4.24 & 65.15$\pm$16.62 & 74.47$\pm$3.86 & 33.26$\pm$8.48 \\
 & \textcolor{red}{IPLC$^{\#}$} & 81.14$\pm$4.24 & 59.92$\pm$6.56 & 74.47$\pm$3.86 & 41.33$\pm$11.27 \\
 & \textcolor{red}{TEGDA$^{\#}$} & 81.14$\pm$4.24 & 66.65$\pm$11.46 & 74.47$\pm$3.86 & 36.30$\pm$6.99 \\
 & \textcolor{green}{Ours$^{\star}$} & \textbf{83.73$\pm$5.47} & \textbf{70.61$\pm$13.27} & \textbf{78.23$\pm$2.94} & \textbf{59.46$\pm$7.03} \\ \hline
\multirow{4}{*}{\begin{tabular}[c]{@{}l@{}}HD95 $\downarrow$\\ (pixel)\end{tabular}}
 & Baseline & 4.84$\pm$2.43 & 17.12$\pm$25.04 & 8.24$\pm$2.78 & 29.97$\pm$13.38 \\
 & \textcolor{red}{IPLC$^{\#}$} & 4.84$\pm$2.43 & 19.35$\pm$28.57 & 8.24$\pm$2.78 & 21.58$\pm$11.86 \\
 & \textcolor{red}{TEGDA$^{\#}$} & 4.84$\pm$2.43 & 14.47$\pm$22.08 & 8.24$\pm$2.78 & 24.39$\pm$13.72 \\
 & \textcolor{green}{Ours$^{\star}$} & \textbf{2.49$\pm$1.99} & \textbf{12.57$\pm$23.31} & \textbf{6.76$\pm$1.83} & \textbf{13.19$\pm$11.07} \\ \hline
\end{tabular}
}
\end{table*}

\begin{figure*}[!t]
\centering
\includegraphics[width=1.0\textwidth]{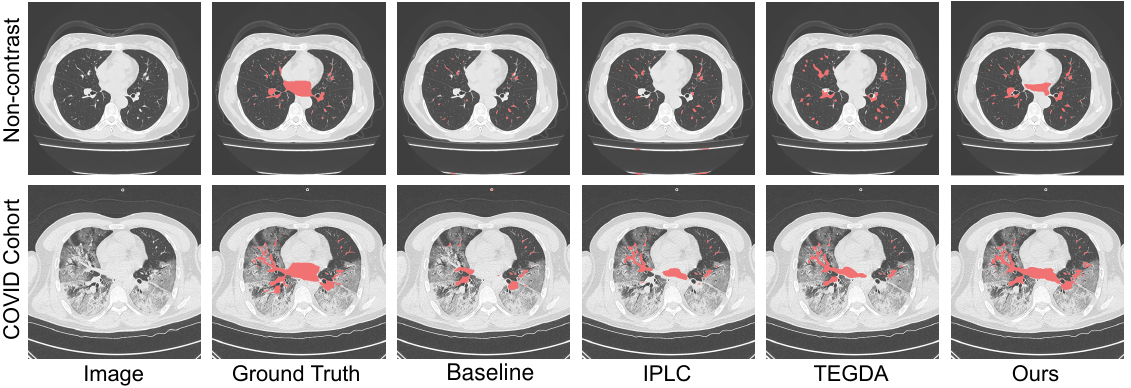}
\caption{Qualitative segmentation results of different methods. The top and bottom rows present the segmentation performance under modality shift (Non-contrast) and demographic shift (COVID Cohort), respectively.}
\label{fig:2}
\end{figure*}

\section{Experiments}
We conducted experiments on one public benchmark and two real-world clinical datasets. These datasets are deliberately chosen because they jointly instantiate the three distribution shifts this work targets---multi-center, modality, and demographic---each posing a distinct generalization challenge.
(1) The M\&Ms dataset~\cite{campello2021multi} contains cardiac MRI scans from four vendors---Siemens (A), Philips (B), GE (C), and Canon (D)---with 192, 252, 150, and 100 volumes, respectively. Domain A served as the source, while B--D were targets, forming a multi-center shift. Segmentation targets include LV, RV, and MYO. Slices were resized to $256 \times 256$. (2) A CT-based lung vessel dataset includes 93 Contrast-enhanced CT and 101 Non-contrast CT volumes, representing a modality shift due to the use of a contrast agent. Contrast-enhanced CT served as the source and Non-contrast CT as the target. Slices were resized to $512 \times 512$. (3) A CT-based lung vessel dataset comprises 98 Normal and 74 COVID-19 cases, reflecting a demographic shift. The Normal Cohort served as the source and the COVID Cohort as the target. The same preprocessing as in (2) was applied. As no public dataset provides COVID-cohort pulmonary vessel annotations, this in-house dataset is required to evaluate the demographic shift. All data collection procedures were approved by the relevant ethics committees, and informed consent was obtained from all participants in accordance with the Declaration of Helsinki.

\noindent\textbf{Implementation Details.} Following previous works~\cite{liu2025spectrum,chen2025dual}, we employ DeepLa\-bv3+~\cite{chen2018encoder} with a MobileNetV2~\cite{sandler2018mobilenetv2} backbone to ensure a fair comparison. All experiments were conducted on a single TITAN RTX GPU with 24\,GB of memory. The models were optimized for 60,000 iterations using Adam (learning rate 0.001), with a batch size of 10 for M\&Ms~\cite{campello2021multi} and 35 for private CT datasets. For CRISP, we set the latent perturbation steps to $N=10$, noise standard deviation to $\sigma=0.35$, confidence grades to $L=5$, and the squeezing loss weight to $\alpha=0.5$. All comparison methods were run with their recommended hyperparameters for a fair comparison. Each reported number comes from a single training run (except the A+T Reference and IPLC, which use 5-fold cross-validation). For reproducibility, all sources of randomness---network initialization, data augmentation, and latent Gaussian perturbation sampling---are controlled by a fixed random seed, so every reported number is exactly reproducible. Performance is evaluated via volume-level Dice and HD95. Throughout the paper, the reported ``$\pm$'' denotes the standard deviation across test volumes (case-level variability) rather than variation across training runs. At inference, CRISP simply repeats a refinement step until the HP/HR hints converge---typically only about three steps. Each step costs roughly two forward passes (one prediction plus one HP/HR update), so the total inference cost is just a small multiple of a single pass (about $6\times$ in our setting), trading modest compute overhead for substantially improved robustness.

\noindent\textbf{Comparison with SOTAs.} On M\&Ms we compare CRISP against four state-of-the-art methods: two domain generalization (DG) models, FedDG~\cite{liu2021feddg} and DDG-Med~\cite{cheng2025dynamic}, and two deployment-time adaptation strategies, the SFDA method IPLC~\cite{zhang2024iplc} and the TTA approach TEGDA~\cite{zhou2025tegda}. We also report two references: Baseline, the source model tested directly on each target; and the A+T Joint Supervised Reference (``A+T Ref.''), a separate model trained on the source $A$ plus each labeled target ($A{+}B$, $A{+}C$, $A{+}D$)---a strong reference rather than a true oracle ceiling, as it still sees only one target (plus the source). Both A+T Ref.\ and IPLC are evaluated by 5-fold cross-validation over the entire target domain; the gap between our reproduced IPLC and its originally reported numbers is mainly a protocol difference---the original study evaluated the method on a held-out 20\% subset, whereas we used the full target under a stricter source-only access setting.

Table~\ref{tab:results} reports per-class Dice and HD95 on the three M\&Ms target domains. Despite being strictly source-only ($\star$), CRISP attains the best Dice in $7$ of the $9$ class--domain cells and the best HD95 in $5$ of $9$, matching or surpassing the two target-informed adaptation methods (IPLC and TEGDA, marked $\#$) that additionally consume target data. Relative to the source-only Baseline, CRISP improves Dice consistently across all domains---e.g., on target $C$ by $+5.60$ (LV), $+5.28$ (MYO), and $+6.16$ (RV), and on target $B$ by up to $+8.24$ (RV)---with the largest gains on the hardest structures (RV and MYO), which the Baseline segments least reliably. Boundary precision improves in tandem: on target $C$, HD95 drops from $3.57$ to $2.06$\,px (LV), $2.66$ to $2.17$\,px (MYO), and $3.94$ to $2.20$\,px (RV). Notably, CRISP even surpasses the A+T Joint Reference in boundary precision across the majority of categories, achieving an overall average HD95 reduction of $0.23$\,px over this jointly-supervised reference---despite using no target data at all. This likely reflects that joint supervised training improves region overlap (Dice) but does not necessarily optimize boundary stability under cross-domain anatomical variability, which is precisely what CRISP's rank-stable refinement targets.

Table~\ref{tab:combined_scenarios} examines two more severe, real-world shifts. Under the modality shift (Contrast-enhanced$\rightarrow$Non-contrast), the Baseline collapses to $33.26$ Dice; the target-informed methods recover only partially (IPLC $41.33$, TEGDA $36.30$), whereas source-only CRISP reaches $59.46$---a $+26.20$ gain over the Baseline and $+18.13$ over the best competitor---and cuts HD95 from $29.97$ to $13.19$\,px ($8.39$\,px below the strongest competitor). Under the demographic shift (Normal$\rightarrow$COVID), CRISP again leads with $70.61$ Dice ($+5.46$ over Baseline, $+3.96$ over the best competitor) and the lowest HD95 ($12.57$\,px). Strikingly, the target-informed methods can degrade performance under such severe shifts---IPLC drops COVID Dice to $59.92$, below the $65.15$ Baseline, and raises its HD95---exposing the fragility of parameter-updating adaptation, whereas CRISP improves reliably without any target data. It even slightly improves performance on the source domains (Normal $+2.59$, Contrast-enhanced $+3.76$ Dice), confirming that hint-conditioned refinement does not sacrifice source performance. The qualitative comparison in Fig.~\ref{fig:2} corroborates these trends: by leveraging inherent anatomical stability rather than chasing distribution shifts, CRISP yields cleaner, better-localized boundaries without any target-domain information.

\begin{figure}[t]
\centering
\includegraphics[width=0.48\linewidth]{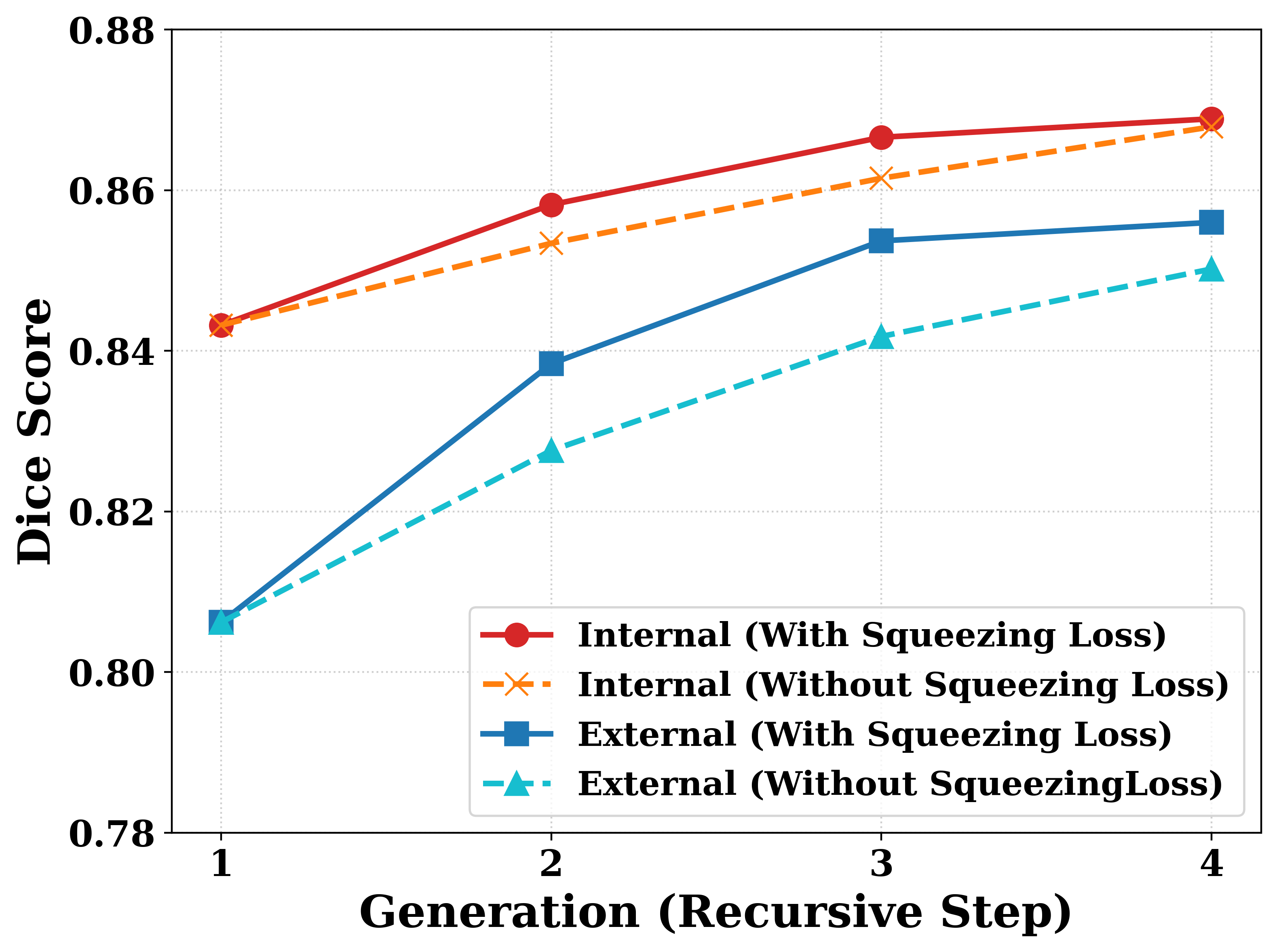}\hfill
\includegraphics[width=0.48\linewidth]{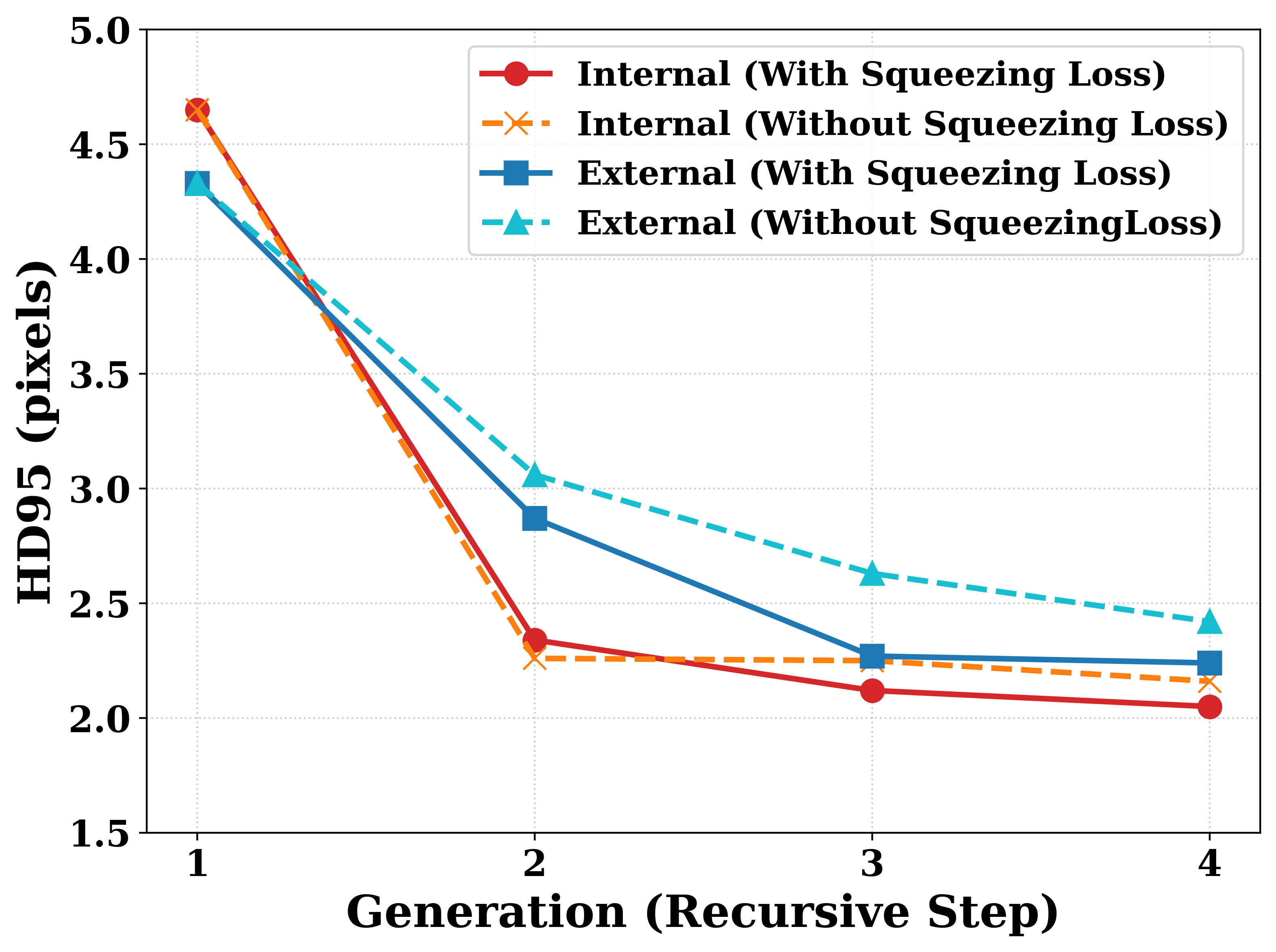}
\caption{Convergence analysis of Dice (left) and HD95 (right) scores.}
\label{fig:ablation1}
\end{figure}

\noindent\textbf{Ablation Study.}
\noindent\emph{Convergence dynamics.} Fig.~\ref{fig:ablation1} tracks Dice and HD95 across refinement generations on the M\&Ms target domains. The Uncertainty Squeezing Loss drives both metrics to converge by the 3rd generation---after which Dice plateaus and HD95 keeps decreasing with shrinking variance---whereas without it convergence is far slower. A few refinement steps $K$ therefore suffice at inference.

\begin{table}[t]
\centering
\caption{Ablation study of CRISP on three distinct challenges.}
\label{tab:ablation}
\setlength{\tabcolsep}{1.5pt}
\resizebox{\columnwidth}{!}{
\begin{tabular}{cccccccc}
\toprule
\multicolumn{1}{c}{Recursive} &
\multicolumn{1}{c}{Uncertainty} &
\multicolumn{2}{c}{M\&Ms} &
\multicolumn{2}{c}{Non-contrast} &
\multicolumn{2}{c}{COVID Cohort} \\
\multicolumn{1}{c}{Self-Evolution} &
\multicolumn{1}{c}{Squeezing Loss} &
Dice$\uparrow$ & HD95$\downarrow$ &
Dice$\uparrow$ & HD95$\downarrow$ &
Dice$\uparrow$ & HD95$\downarrow$ \\
\midrule
 &  &
 \makecell{80.62 \\ $\pm$2.54} & \makecell{4.33 \\ $\pm$1.62} &
 \makecell{33.26 \\ $\pm$8.48} & \makecell{29.97 \\ $\pm$13.38} &
 \makecell{65.15 \\ $\pm$16.62} & \makecell{17.12 \\ $\pm$25.04} \\
\ding{51} &  &
 \makecell{85.02 \\ $\pm$0.56} & \makecell{2.42 \\ $\pm$0.13} &
 \makecell{51.45 \\ $\pm$6.57} & \makecell{19.80 \\ $\pm$12.86} &
 \makecell{67.42 \\ $\pm$12.48} & \makecell{16.27 \\ $\pm$23.53} \\
\ding{51} & \ding{51} &
\makecell{\textbf{85.60} \\ \textbf{$\pm$0.85}} &
\makecell{\textbf{2.24} \\ \textbf{$\pm$0.42}} &
\makecell{\textbf{59.46} \\ \textbf{$\pm$7.03}} &
\makecell{\textbf{13.19} \\ \textbf{$\pm$11.07}} &
\makecell{\textbf{70.61} \\ \textbf{$\pm$13.27}} &
\makecell{\textbf{12.57} \\ \textbf{$\pm$23.31}} \\
\bottomrule
\end{tabular}
}
\end{table}
\noindent\emph{Component analysis.} Table~\ref{tab:ablation} isolates the two components across all three shifts: a baseline without adaptation, Recursive Self-Evolution (RSE) alone, and the full CRISP. RSE supplies the perturbation-derived spatial hints and contributes the bulk of the improvement. In Dice it lifts the baseline by $+4.40$ on the multi-center (M\&Ms) shift, by a dramatic $+18.19$ on the modality (Non-contrast) shift, and by $+2.27$ on the demographic (COVID) shift; the same trend holds for boundary quality, with HD95 falling by $1.91$, $10.17$, and $0.85$\,px on the three shifts, respectively. Equally important, RSE sharply stabilizes the predictions---on M\&Ms the Dice std collapses from $2.54$ to $0.56$ and the HD95 std from $1.62$ to $0.13$---removing most of the catastrophic per-case failures that plague the unadapted baseline. The Uncertainty Squeezing Loss then adds a further Dice gain of $+0.58$ (M\&Ms), $+8.01$ (Non-contrast), and $+3.19$ (COVID), and reduces HD95 by an additional $0.18$, $6.61$, and $3.70$\,px (to $2.24$, $13.19$, and $12.57$\,px); its effect is concentrated on boundary precision and grows with shift severity---largest on the most degraded Non-contrast and smallest on the already-saturated M\&Ms---exactly as argued in Sec.~\ref{sec:2.3}: when the foreground evidence inside $\Omega_u$ is weakest and the rank signal most fragile, enforcing perturbation-invariance contributes the most. The two address different failure modes---RSE injects structural priors that recover gross region overlap, whereas the squeezing loss tightens the residual ambiguous band into a sharp, stable boundary---so their gains are additive: end-to-end CRISP improves Dice by up to $+26.20$ (Non-contrast) and cuts HD95 by $16.78$\,px, confirming the two are complementary rather than redundant.

\noindent\emph{Worst-case robustness.} We further analyze the full per-case Dice distribution on the Non-contrast target ($N{=}101$), pairing the Baseline and CRISP results for each volume. The gains concentrate in the tail: while the mean Dice rises by $0.26$, the average Dice score among the worst $5\%$ of cases improves from $0.175$ to $0.433$ and the single worst case from $0.07$ to $0.33$. Critically, $36.4\%$ of Baseline volumes are near-total failures (Dice$<0.30$), a regime CRISP eliminates entirely ($0\%$), and the share of failed segmentations (Dice$<0.50$) collapses from $97.4\%$ to $6.5\%$. CRISP therefore does not merely lift average accuracy---it removes the long tail of catastrophic, clinically unusable segmentations, which is precisely where deployment risk resides.

\section{Conclusion}
We proposed CRISP, a model-agnostic, source-only framework for medical image segmentation that requires no test-time parameter updates and no target-domain data. Built on the Rank Stability of Positive Regions, CRISP uses Latent Feature Perturbation to derive High-Precision and High-Recall priors and recursively squeezes them, guided by the Uncertainty-Squeezing Loss, toward the final segmentation. Across multi-center, modality, and demographic shifts on cardiac MRI and CT-based lung vessel segmentation, CRISP outperforms state-of-the-art methods.

Crucially, this advantage requires no access to the target domain, yet CRISP matches or surpasses target-informed adaptation methods without incurring their risk of negative adaptation. The gains are largest exactly where the gap is widest---on the severe non-contrast shift, CRISP improves the collapsed baseline from a Dice score of $33.26$ to $59.46$, and the boundary error is nearly halved (from $29.97$ to $13.19$\,px). Because these gains concentrate in the tail and remove the catastrophic, near-zero-overlap failures that govern clinical deployment risk, CRISP degrades gracefully where existing methods fail abruptly.

Future work will explore unrolling the recursive process into a unified, stochastic training loop. Training on mixed-stage batches enables the model to learn a universal squeezing policy for efficient self-convergence.

\bibliography{crispref}

\end{document}